\documentclass[10pt,twocolumn,letterpaper]{article}

\usepackage{cvpr}              % To produce the CAMERA-READY version
\usepackage{graphicx}
\usepackage{amsmath}
\usepackage{amsthm}
\usepackage{amssymb}
\usepackage{booktabs}
\usepackage{enumitem}
\usepackage[dvipsnames, table]{xcolor}
\usepackage{pifont}
\usepackage{multirow}
\usepackage{makecell}
\usepackage{paralist}
\usepackage{array}
\usepackage[accsupp]{axessibility}

\newcommand{\cmark}{\ding{51}}
\newcommand{\xmark}{\ding{55}}

\newcommand*\samethanks[1][\value{footnote}]{\footnotemark[#1]}

\usepackage[pagebackref,breaklinks,colorlinks]{hyperref}

\usepackage{flushend}

\usepackage[capitalize]{cleveref}
\crefname{section}{Sec.}{Secs.}
\Crefname{section}{Section}{Sections}
\Crefname{table}{Table}{Tables}
\crefname{table}{Tab.}{Tabs.}

\def\cvprPaperID{6072} % *** Enter the CVPR Paper ID here
\def\confName{CVPR}
\def\confYear{2023}

\begin{document}

%%%%%%%%% TITLE - PLEASE UPDATE
\title{Learning Emotion Representations from Verbal and Nonverbal Communication}

\author{Sitao Zhang\thanks{equal contribution} \qquad Yimu Pan\samethanks \qquad James Z. Wang\\
The Pennsylvania State University, University Park, Pennsylvania, USA\\
{\tt\small \{sbz5211,ymp5078,jwang\}@psu.edu}
% For a paper whose authors are all at the same institution,
% omit the following lines up until the closing ``}''.
% Additional authors and addresses can be added with ``\and'',
% just like the second author.
% To save space, use either the email address or home page, not both
% \and
% Second Author\\
% Institution2\\
% First line of institution2 address\\
% {\tt\small secondauthor@i2.org}
}
\maketitle

%%%%%%%%% ABSTRACT
\begin{abstract}
Emotion understanding is an essential but highly challenging component of artificial general intelligence. The absence of extensively annotated datasets has significantly impeded advancements in this field. We present EmotionCLIP, the first pre-training paradigm to extract visual emotion representations from verbal and nonverbal communication using only uncurated data. Compared to numerical labels or descriptions used in previous methods, communication naturally contains emotion information. Furthermore, acquiring emotion representations from communication is more congruent with the human learning process. We guide EmotionCLIP to attend to nonverbal emotion cues through subject-aware context encoding and verbal emotion cues using sentiment-guided contrastive learning. Extensive experiments validate the effectiveness and transferability of EmotionCLIP. Using merely linear-probe evaluation protocol, EmotionCLIP outperforms the state-of-the-art supervised visual emotion recognition methods and rivals many multimodal approaches across various benchmarks. We anticipate that the advent of EmotionCLIP will address the prevailing issue of data scarcity in emotion understanding, thereby fostering progress in related domains. The code and pre-trained models are available at \href{https://github.com/Xeaver/EmotionCLIP}{https://github.com/Xeaver/EmotionCLIP}.
\end{abstract}

%%%%%%%%% BODY TEXT
\section{Introduction}
\label{sec:intro}
% outline:
% importance of Emotion Intelligence
% lack of progress in emotion understanding
% current work is not suitable
% how human learns -> our approach
% summary of contributions
\begin{figure}[ht!]
    \centering
    \includegraphics[width=0.95\linewidth]{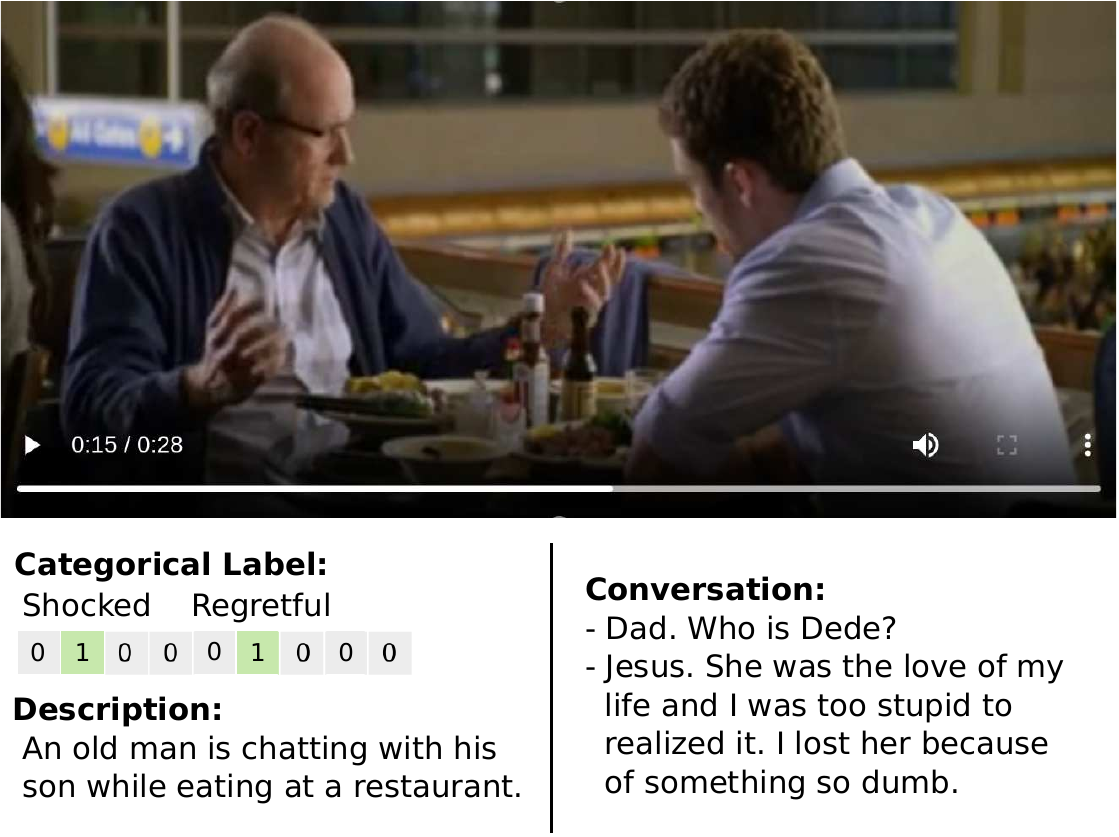}
    \caption{Emotions emerge naturally in human communication through verbal and nonverbal cues. The rich semantic details within the expression can hardly be represented by human-annotated categorical labels and descriptions in current datasets.}
    \label{fig:motivation}
\end{figure}

If artificial intelligence (AI) can be equipped with emotional intelligence (EQ), it will be a significant step toward developing the next generation of artificial general intelligence~\cite{yonck2020heart, maruyama2020agiconditions}. The combination of emotion and intelligence distinguishes humans from other animals. The ability to understand, use, and express emotions will significantly facilitate the interaction of AI with humans and the environment~\cite{mehrabian1980basic,evans_2020, mehrabian1972nonverbal, mehrabian1971silent}, making it the foundation for a wide variety of HCI~\cite{bartneck2020human}, robotics~\cite{christensen2021roadmap}, and autonomous driving~\cite{kraus2009cognition} applications. 

Artificial emotional intelligence (AEI) research is still in its nascency~\cite{schuller2018age, krakovsky2018aei}. The recent emergence of pre-trained models in CV~\cite{chen2020simclr, he2020moco, CLIP} and NLP~\cite{devlin2018bert, sanh2019distilbert, brown2020gpt, lample2019xlm} domains has ushered in a new era of research in related subjects. By training on large-scale unlabeled data in a self-supervised manner, the model learns nontrivial representations that generalize to downstream tasks~\cite{liu2019roberta, CLIP}. Unfortunately, such a technique remains absent from AEI research. The conventional approaches in visual emotion understanding have no choice but to train models from scratch, or leverage models from less-relevant domains~\cite{kay2017kinetics, russakovsky2015imagenet}, suffering from data scarcity~\cite{emotic, BoLD}. The lack of pre-trained models greatly limits the development of AEI research.

Research in neuroscience and psychology offers insights for addressing this problem. Extending from the capabilities that have been coded genetically, humans learn emotional expressions through daily interaction and communication as early as when they are infants. It has been shown that both vision~\cite{pessoa2010emotion} and language~\cite{lindquist2015role} play crucial roles in this learning process. By absorbing and imitating expressions from others, humans eventually master the necessary feelings to comprehend emotional states by observing and analyzing facial expressions, body movements, contextual environments, \etc.

Inspired by how humans comprehend emotions, we propose a new paradigm for emotion understanding that learn directly from human communication. The core of our idea is to explore the consistency between verbal and nonverbal affective cues in daily communication. Fig.~\ref{fig:motivation} shows how communication reveals emotion. Our method that learns from communication is not only aligned with the human learning process but also has several advantages:

1)
{\it Our method bypasses the problems in emotion data collection by leveraging uncurated data from daily communication.} 
Existing emotion understanding datasets are mainly annotated using crowdsourcing~\cite{emotic, BoLD, moviegraphs}. For image classification tasks, it is straightforward for annotators to agree on an image's label due to the fact that the label is determined by certain low-level visual characteristics. However, crowdsourcing participants usually have lower consensus on producing emotion annotations due to the subjectivity and subtlety of affective labels~\cite{ye2017probabilistic}. This phenomenon makes it extremely difficult to collect accurate emotion annotations on a large scale. Our approach does not rely on human annotations, allowing us to benefit from nearly unlimited web data.

2)
{\it Our use of verbal expressions preserves fine-grained semantics to the greatest extent possible.} 
Limited by the data collection strategy, existing datasets usually only contain annotations for a limited number of emotion categories, which is far from covering the space of human emotional expression~\cite{wortman2022hicem}. Moreover, the categorical labels commonly used in existing datasets fail to precisely represent the magnitude or intensity of a certain emotion.

3)
{\it Our approach provides a way to directly model expressed emotion.} Ideally, AEI should identify the individual's emotional state, i.e., the emotion the person desires to express. Unfortunately, it is nearly impossible to collect data on this type of ``expressed emotion'' on a large scale. Instead, the current practice is to collect data on ``perceived emotion'' to approximate the person's actual emotional state, which inevitably introduces noise and bias to labels. 

In general, learning directly from how humans express themselves is a promising alternative that gives a far broader source of supervision and a more comprehensive representation. This strategy is closely analogous to the human learning process and provides an efficient solution for extracting emotion representations from uncurated data.

We summarize our main {\bf contributions} as follows:
\begin{compactitem}
\item We introduce EmotionCLIP, the first vision-language pre-training paradigm using uncurated data to the visual emotion understanding domain.
\item We propose two techniques to guide the model to capture salient emotional expressions from human verbal and nonverbal communication.
\item Extensive experiments and analysis demonstrate the superiority and transferability of our method on various downstream datasets in emotion understanding.
\end{compactitem}

\section{Related Work}

\noindent{\bf Emotion Recognition from Visual Clues.}
Facial expression recognition~\cite{facial_survey_new, facial_survey_old, facs} has been well studied in the field of emotion recognition, mainly because faces are not only expressive but also easy to model. Handcrafted features have been developed to describe different facial expressions~\cite{lowe2004sift, dalal2005hog, shan2009lbp}. Recently, deep learning-based approaches have begun to emerge~\cite{facial_survey_new}. Principal research focuses on the design of novel modules for conventional network architectures~\cite{xue2021transfer}, distinct loss functions for facial tasks~\cite{deng2019arcface, schroff2015facenet, wen2016centerloss}, and addressing label uncertainties~\cite{wang2020suppressing, chen2020label}.
% Another line of research is based on the Facial Action Coding System (FACS), which systematically encodes the movements of individual facial muscles in facial appearance.

Recently, with the growing interest in recognizing emotion in the wild, the focus of research has gradually shifted to modeling body language~\cite{BoLD, Filntisis2020bold, Pikoulis2021LeveragingSS} and context~\cite{emotic, affect2mm, emoticon}. Several datasets for understanding human emotional states in unconstrained environments have been proposed~\cite{moviegraphs, Liris-Accede, MELD}; Kosti~\etal~\cite{emotic} and Yu~\etal~\cite{BoLD} established the first benchmark for image and video data, respectively.
Follow-up work mainly focuses on context-aware emotion recognition, which usually adopts a multi-branch structure where one branch focuses on the face or body and the other focuses on capturing context~\cite{CAER, emoticon, m2fnet, Pikoulis2021LeveragingSS}. Moreover, some approaches take into account temporal causality~\cite{affect2mm} or represent context information via graphs~\cite{affective_graph}. To the best of our knowledge, there are no pre-trained models or effective methods for leveraging unlabeled data in the domain of visual emotion recognition.

\begin{figure*}[ht!] 
    \centering
   \includegraphics[width=0.9\linewidth]{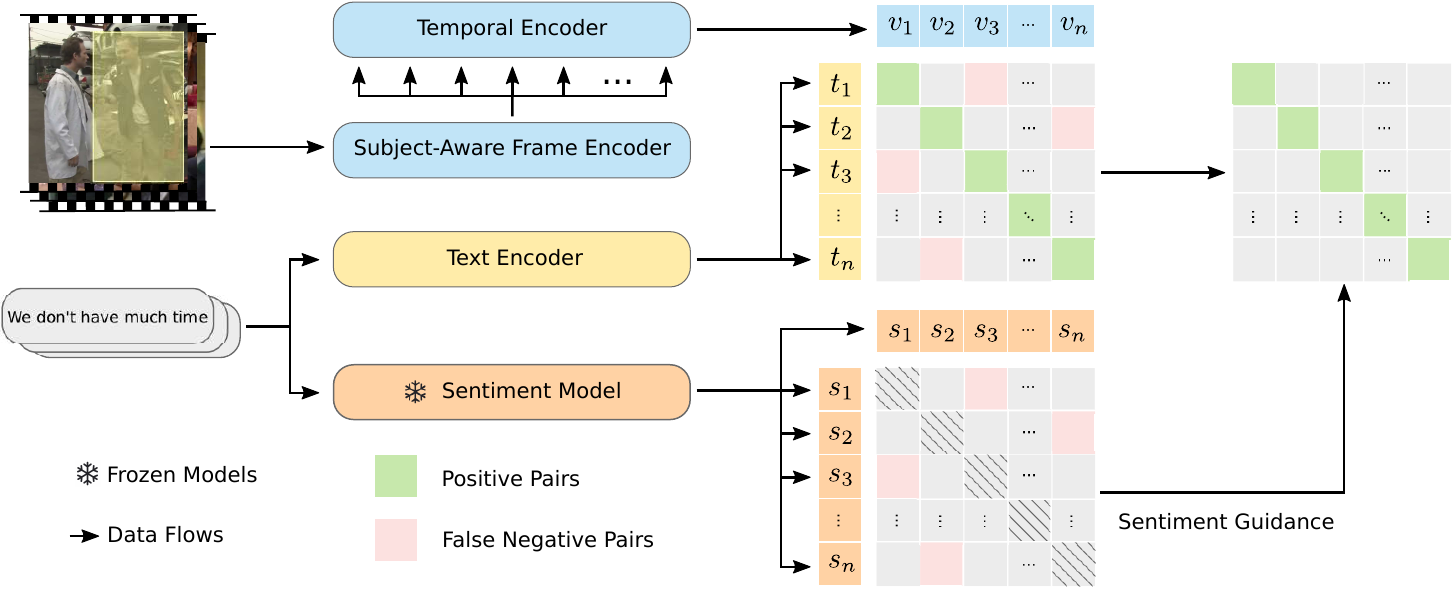}
    % \caption{The illustration of EmotionCLIP. The subject-aware encoder maps the frames and the subject positions into frame features. The temporal encoder aggregates the frame features into video features. The text encoder and a pre-trained sentiment model map the input text into text features and sentiment scores. The similarity of sentiment scores is used to reweight the negative samples in contrastive loss.}
    \caption{Illustration of \textbf{EmotionCLIP}. For \textit{nonverbal communication}, subject and context information is modeled by a frame encoder and further aggregated into video-level representations by a temporal encoder. For \textit{verbal communication}, textual information is encoded as text representations and sentiment scores by a text encoder and sentiment analysis model, respectively. The model learns emotion representations under sentiment guidance in a contrastive manner, by exploring the consistency of verbal and nonverbal communication.}
    \label{fig:model_overview}
\end{figure*}

\noindent{\bf Vision-Language Pre-training.}
Visual-language pre-training has achieved remarkable progress recently. CLIP~\cite{CLIP} demonstrated the feasibility of using contrastive learning~\cite{chen2020simclr, oord2018nce} to learn transferable and powerful visual representations from large-scale image-text pairs~\cite{schuhmann2021laion}. Many follow-up approaches have been proposed to transfer the pre-trained model to downstream tasks~\cite{zhou2022coop, gao2021adapt, zhang2021tip} or leverage the scheme for different domains~\cite{wei2022hairclip, rao2022denseclip, zhang2022pointclip, li2022glip, yimu2022miccai}. A line of research endeavors to expand CLIP for general video understanding~\cite{luo2022clip4clip, lin2022frozen, ju2021prompting, videoclip, xclip, wang2021actionclip, lei2021clipbert}. The majority of the effort focuses on fine-tuning datasets with textual annotations~\cite{kay2017kinetics, lin2014mscoco, krishna2017visualgenome}. However, not only are these curated annotations challenging to obtain, but they also limit the model's potential in various applications. Another line of work~\cite{li2020hero, videoclip} extends the image-level pre-training by utilizing unlabeled narrated videos~\cite{miech2019howto100m, webvid10m}, similar to ours. However, we aim to learn abstract {\it emotion representations} rather than low-level visual patterns, which are beyond the reach of current models. 
EmotionNet~\cite{emotionnet} and its sequel\cite{ijcai2022emotion}, which likewise seeks to learn visual emotion representations, are connected to ours. However, their primary focus is on the images' stimuli rather than the recognition of human emotional expressions in the wild.

% \section{Datasets}

% It is generally believed that people who utter similar phrases and display comparable behaviors under similar circumstances tend to exhibit and provoke similar feelings and that these emotions must be present in the scenario in terms of the contextual background, facial expression, behavior, or body language. This type of pre-text task is more akin to emotion comprehension than other pre-training tasks using publicly accessible datasets, as the suggested dataset infers emotion information whereas the existing datasets do not. If a relationship between closed captions and video clips can be recorded using the VLCP architecture, we will need considerably fewer distracting emotion labels to learn appropriate emotion representations.

\section{Methodology}
\label{sec:method}

Our core idea is to learn directly from human communication how they express their emotions, by exploring the consistency between their verbal and nonverbal expressions~\cite{lindquist2015role}. We tackle this learning task under the vision-language contrastive learning framework~\cite{CLIP}, that is, the model is expected to learn consistent emotion representations from the verbal expressions (\eg, utterance and dialogue) and nonverbal expressions (\eg, facial expression, body language, and contextual environment) of the same individuals. 
We give a brief introduction of our data collection procedure in Sec.~\ref{sec:data_collection} before presenting the overview of EmotionCLIP in Sec.~\ref{sec:method_overview}. We further discuss how the model is guided to learn emotion-relevant representations from the nonverbal perspective in Sec.~\ref{sec:method_subject}, and from the verbal perspective in Sec.~\ref{sec:method_reweighting}. Please see Appendix for details of the dataset and implementations.

\subsection{Data Collection}
\label{sec:data_collection}
Publicly available large-scale vision-and-language datasets do not provide desired verbal and nonverbal information because they either comprise only captions of low-level visual elements~\cite{schuhmann2021laion, webvid10m} or instructions of actions~\cite{miech2019howto100m}. 
The captions mostly contain a brief description of the scene or activity, which is insufficient to reveal the underlying emotions; the instruction videos rarely include humans in the scene or express neutral emotions, which fail to provide supervision signals for emotion understanding.
To overcome such problems, we gather a large-scale video-and-text paired dataset. More specifically, the videos are TV series, while the texts are the corresponding closed captions. We collected 3,613 TV series from YouTube, which is equivalent to around a million raw video clips. We processed them using the off-the-shelf models to group the words in closed caption into complete sentences~\cite{guhr-EtAl:2021:fullstop}, tag each sentence with a sentiment score~\cite{sanh2019distilbert}, and extract human bounding boxes~\cite{wang2022yolov7}.

\subsection{Overview of EmotionCLIP}
\label{sec:method_overview}
Fig.~\ref{fig:model_overview} presents an overview of our approach. We follow the wildly adopted vision-language contrastive learning paradigm~\cite{CLIP} where two separate branches are used to encode visual inputs (\ie, nonverbal expressions) and textual inputs (\ie, verbal expressions), respectively.

\noindent{\bf Video Encoding.}
The visual branch of EmotionCLIP takes two inputs, including a sequence of RGB frames $X_v$ and a sequence of binary masks $X_m$. The binary mask has the same shape as the frame and corresponds to the frame one-to-one, indicating the location of the subject within the frame. The backbone of the subject-aware frame encoder $f_i$ is a Vision Transformer~\cite{dosovitskiy2020vit}. In particular, it extracts $m$ non-overlapping image patches from the frame and projects them into 1D tokens $z_i \in \mathbb{R}^d$. The sequence of tokens passed to the following Transformer encoder~\cite{vaswani2017attention} is $\mathbf{z} = [z_1, \cdots, z_m, z_{cls}, z_{hmn}]$,
% \begin{equation}
%     \mathbf{z} = [z_1, \cdots, z_m, z_{cls}, z_{hmn}],
% \end{equation}
where $z_{cls}, z_{hmn}$ are two additional learnable tokens. The mask $X_m$ is converted to an array of indices $P$ indicating the image patches containing the subject.
The frame encoder further encodes $\mathbf{z}, P$ into a frame-level representation. All frame representations are then passed into the temporal encoder $f_p$ to produce a video-level representation as $v = f_v(\mathbf{z}, P)$, where $f_v = f_p \circ f_i$ and $v\in \mathbb{R}^d$.
% \begin{equation}
%     v = f_v(\mathbf{z}, P), \quad f_v = f_p \circ f_i
% \end{equation}

\noindent{\bf Text Encoding.}
The textual branch of EmotionCLIP takes sentences $X_t$ as inputs. The text encoder $f_t$ is a Transformer~\cite{devlin2018bert} with the architecture modification described in~\cite{radford2019bertrefined}, and the sentiment model $f_s$ is a pre-trained sentiment analysis model~\cite{sanh2019distilbert} that is frozen during training. The input text is encoded by both models as $t = f_t(X_t)$ and $s = f_s(X_t)$,
% \begin{equation}
%     t = f_t(X_t), \quad s = f_s(X_t),
% \end{equation}
where $t \in \mathbb{R}^d$ is the representation of the text and $s \in \mathbb{R}^7$ is the pseudo sentiment score.

\noindent{\bf Training Objective.}
The training objective is to learn the correspondence between visual inputs and textual inputs by minimizing the sentiment-guided contrastive loss $\mathcal{L}$.
% \begin{equation}
%     \mathcal{L} = \sum_{(v,t,s)\in B} \bigl(\text{InfoNCE}^*(v, t, s) + \text{InfoNCE}^*(t, v, s)\bigr),
% \end{equation}
% where $B$ is a batch, $v, t, s$ are the outputs of temporal encoder, text encoder and sentiment model respectively.

\noindent We discuss the details of the proposed subject-aware encoding approaches in Sec.~\ref{sec:method_subject} and the sentiment-guided contrastive learning framework in Sec.~\ref{sec:method_reweighting}.

\subsection{Subject-Aware Context Encoding}
\label{sec:method_subject}
\begin{figure}[t] 
    \centering
    \includegraphics[width=\linewidth]{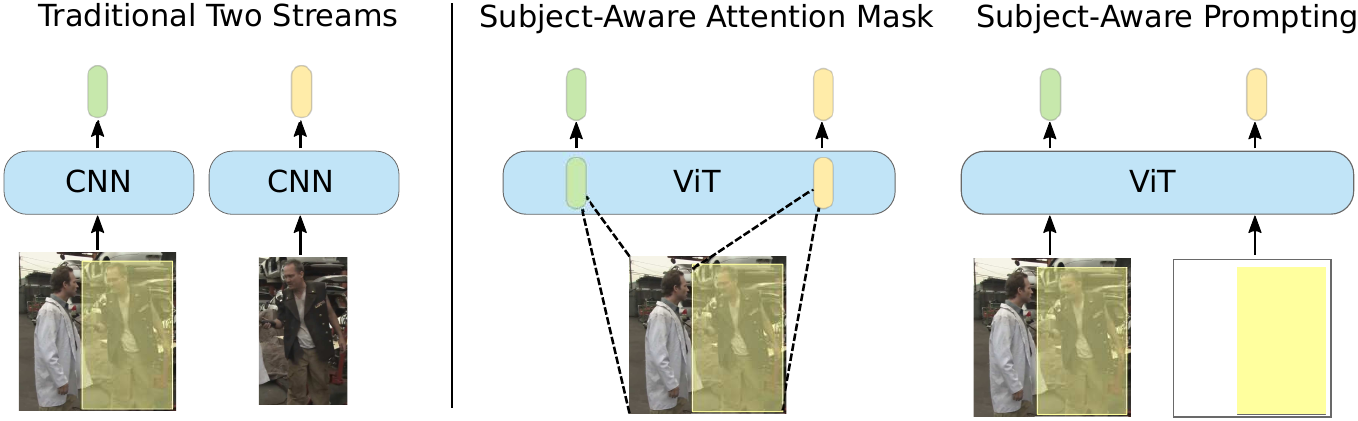}
    \caption{The traditional approaches ({\it left}) ignore the dependencies between context and subject, and encode a portion of the image redundantly. The proposed approaches ({\it right two diagrams}) efficiently model the subject and context in a synchronous way.}
    \label{fig:subject_aware}
\end{figure}

Context encoding is an important part of emotion understanding, especially in unconstrained environments, as it has been widely shown in psychology that emotional processes cannot be interpreted without context~\cite{frege, mcnulty2012beyond, mesquita2010mind}. We intend to guide the model to focus on the interaction between the subject of interest and context.
% Previous methods fail to provide acceptable solutions to this problem. 
As shown in Fig.~\ref{fig:subject_aware}, the cropped character and the whole image are usually encoded by two separate networks and fused at the ends~\cite{CAER, emoticon, Pikoulis2021LeveragingSS}. This approach is inflexible and inefficient since it overlooks the dependency between subject and context and encodes redundant image portions.
% On the one hand, it ignores the dependencies of context and subject, modeling the two completely independently. On the other hand, redundant encoding is performed on a portion of the image (the subject's region). This deficiency is particularly serious when there are multiple characters in the scene, as this modeling strategy fails to capture the subject of interest in context encoding. 
Following this line of thought, we propose two potential subject-aware context encoding strategies, \ie, subject-aware attention masking (SAAM) and subject-aware prompting (SAP). 
The former can be regarded as an efficient implementation of the traditional two-stream approach but avoids the problem of redundant encoding. The latter is a novel encoding strategy that enables adaptive modeling of the interaction between the context and the subject by providing necessary prompts.

\paragraph{Subject-Aware Attention Masking.}
\label{sec:method_attnmask}
The canonical attention module~\cite{vaswani2017attention} in a Transformer is defined as:
\begin{equation}\label{eq:attention}
    \mathrm{Attention}(\mathbf{Q}, \mathbf{K}, \mathbf{V}) = \mathrm{softmax}\left(\frac{\mathbf{Q}\mathbf{K}^\top}{\sqrt{d}}\right)\mathbf{V}\;.
\end{equation}

We model the context and subject in a synchronous way by modifying the attention module to
\begin{equation}\label{eq:attention_masked}
\begin{split}
&\mathrm{Attention}^*(\mathbf{Q}, \mathbf{K}, \mathbf{V}, \mathbf{U}) = \\ 
&\underbrace{\mathrm{softmax}\left(\frac{\mathbf{Q}\mathbf{K}^\top}{\sqrt{d}}\right)(\mathbf{J}-\mathbf{A})\mathbf{V}}_{\text{\small context}} +
\underbrace{\mathrm{softmax}\left(\frac{\mathbf{Q}\mathbf{K}^\top}{\sqrt{d}}\right)\mathbf{A}\mathbf{U}\mathbf{V}}_{\text{\small subject}}\;,
\end{split}
\end{equation}
where $\mathbf{J}$ is a matrix with all ones, $\mathbf{A}$ is a learnable parameters containing values in range $[0,1]$, and $\mathbf{U}$ is a weight matrix constructed using $P$. Intuitively, we shift $\mathbf{A}$ amount of attention from a total of $\mathbf{J}$ amount of attention from the context to the subject. To partition the $\mathbf{A}$ amount of attention to all image patches containing subject, we compute $\mathbf{U}$ as the following:
\begin{equation}\label{eq:attention_softmax}
    \begin{split}
    \mathbf{U} = \mathrm{softmax}\left(\frac{\mathbf{Q}\mathbf{K}^\top+\mathbf{M}}{\sqrt{d}}\right)\;.
    \end{split}
\end{equation}
The masking matrix $\mathbf{M}$ is defined as:
\begin{equation} \label{eq:con_loss_reweighting}
\mathbf{M}=
\begin{bmatrix}
\mathbf{M}^{(1)} & \mathbf{M}^{(2)}\\
\mathbf{M}^{(3)} & \mathbf{M}^{(4)}\\ 
\end{bmatrix}\;,
\end{equation}
where $\mathbf{M}^{(1)}=\mathbf{0}_{(m+1)\times (m+1)}$, $\mathbf{M}^{(4)}=\mathbf{0}_{1\times 1}$, $\mathbf{M}^{(2)}_{i\notin P}=\mathbf{M}^{(3)}_{i\notin P}=-\infty$, $\mathbf{M}^{(3)}_{(m+1)}=-\infty$, and all other entries are zero. Intuitively, $\mathbf{M}^{(2)}$ and $\mathbf{M}^{(3)}$ represent the attention between all image patches $z_i$ and the human token $z_{hmn}$ to model the subject stream; we mask out all attention between non-human patches and the human token. Moreover, we mask out attention from $z_{hmn}$ to $z_{cls}$,  $\mathbf{M}^{(3)}_{(m+1)}$, to ensure $z_{hmn}$ only encodes the subject.

% This approach satisfies property (1) since the attention is always based on the context and satisfies property (2) since $M$ can be different for the different persons of interest.

\paragraph{Subject-Aware Prompting.}
\label{sec:method_prompting}
Prompting is a parameter-free method that restricts the output space of the model by shaping inputs. In our case, we hope to prompt the model to distinguish between the context and the subject. A recent visual prompting method, CPT~\cite{yao2021cpt}, provides such a prompt by altering the original image, \ie, imposing colored boxes on objects of interest. It shows that a Transformer is able to locate objects with the help of positional hints. However, introducing artifacts on pixel space may not be optimal as it causes large domain shifts. To address this issue, we propose to construct prompts in the latent space based on positional embeddings, considering that they are inherently designed as indicative information.
Formally, let $e_i$ be the positional embedding corresponding to the patch token $z_i$, and $P$ are the indicator set of the subject location. The prompting token is designed as $z_{hmn}=\sum_{i\in P}{e_i}$.
% \begin{equation}
%     z_{hmn}=\sum_{i\in P}{e_i}.
% \end{equation}

We argue the sum of positional embeddings is enough to provide hints about the subject location. The previous study~\cite{wang2020position} demonstrates a Transformer treats all tokens without positional embedding uniformly but with positional embedding differently. This result shows positional embeddings play a vital role in guiding model attention.
% In other words, the transformer must extract positional embedding, $e_i$, from $e_i+z_i$. Since $z_i$ is a random input vector, we expect the same transformer to extract positional embedding, $e_i$, information from $z_{hmn}=e_i+\sum_{j\ne i}{e_j}$. 
  
% Since the vectors in $E$ have different directions, a neural network is able to match the sum of positional embedding to the position of interest unless two different subsets of $E$ happen to sum to the same vector. This case is unlikely to happen since if we view each subset of $E$ as one sample, due to the curse of dimensional, we expect the samples to be sparse.
% $$\mathrm{sim}(e_p+v,T)>\mathrm{sim}(e_{-p}+v,T)$$, where $\mathrm{sim}$ is the cosine similarity and $-p$ is any value that is not in $P$. In other words, a function is able to learn the tokens of interest providing $T$ and $E$. 

% In the convergence, this approach satisfies property (1) because adding $z_{hmn}$ does not block the interaction between other tokens. It also satisfies property (2) because if we change the index in $P$, $z_{hmn}$ changes which affect the final output.

\subsection{Sentiment-Guided Contrastive Learning}
\label{sec:method_reweighting}

We train the model to learn emotion representations from verbal and nonverbal expressions in a contrastive manner. In the traditional contrastive setting, the model is forced to repel all negative pairs except the only positive one. However, many expressions in daily communication indeed have the same semantics from an emotional perspective. Contrasting these undesirable negative pairs encourages the model to learn spurious relations. This problem comes from false negatives, \ie, the affectively similar samples are treated as negatives. We address this issue by introducing a trained sentiment analysis model~\cite{sanh2019distilbert} from the NLP domain for the suppression of false negatives, thereby guiding our model to capture emotion-related concepts from verbal expressions. Specifically, we propose a sentiment-guided contrastive loss:
% We borrow ideas from knowledge distillation by introducing an off-the-shelf analysis model; 
% We tag each text input $X_t$ with a sentiment feature $s$ from an off-the-shelf model and weight each negative sample using
% \begin{equation} \label{eq:con_loss_reweighting}
% w_{t_i,t_j}=
% \begin{cases}
% \beta \cdot \mathrm{KL}(s^{i}\|s^{j})^{-1} & i\ne j \\
% % \frac{\beta}{\mathrm{KL}(s_{i}\|s_{j})}  & i\ne j \\
% 0 & i=j
% \end{cases}
% \end{equation}
% where $\beta$ is a hyper-parameter. 
% Then, $\mathrm{InfoNCE}$ loss becomes
% \begin{equation} \label{eq:con_loss}
% \mathrm{InfoNCE}^*(v,t,s)=
% -\log\frac{\exp(v \cdot t^+ / \tau)}
% {\sum_{t'\in \{t^+, t^-\}} \exp(v \cdot t' / \tau - w_{t^+,t'})}
% \end{equation}.
% Note that when $i\ne j$ but $s_{i}= s_{j}$, we have $w_{t_i,t_j}=\infty$, which is equivalent to removing $t_{j}$ from the negative samples; we set $w_{t_i,t_i}=0$ so that the positive pair is not affected. Since $s$ is the sentiment feature, the sentiment-related difference is weighted more. In other words, we expect the proposed loss to also provide a cleaner supervision signal for emotion-related representations.
\begin{equation} \label{eq:con_loss}
\mathrm{SNCE}(v,t,s)=
-\sum_{i \in B}\left(\log\frac{\exp{(v_i \cdot t_i / \tau)}}
{\sum\limits_{j\in B} \exp{(v_i \cdot t_j / \tau - w_{i,j})}}\right),
\end{equation} 
where $B$ is a batch. The reweighting term $w_{i,j}$ is defined as
\begin{equation} \label{eq:con_loss_w}
w_{i,j}=
\begin{cases}
\beta \cdot \mathrm{KL}(s_{i}\|s_{j})^{-1} & i\ne j \\
% \frac{\beta}{\mathrm{KL}(s_{i}\|s_{j})}  & i\ne j \\
0 & i=j
\end{cases}\;,
\end{equation}
where $\beta$ is a hyper-parameter for controlling the reweighting strength. 
The total loss is defined as:
\begin{equation} \label{eq:final_loss}
\mathcal{L} = \frac{1}{2|B|}\Bigl(\mathrm{SNCE}(v,t,s) + \mathrm{SNCE}(t,v,s)\Bigr)\;.
\end{equation} 
% -\frac{1}{|B|}\sum_{i=0}^{|B|}\left(

\begin{figure}[t] 
    \centering
    \includegraphics[width=0.8\linewidth]{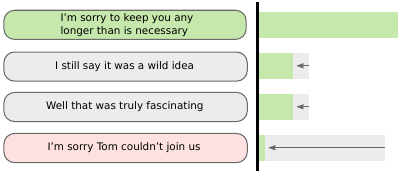}
    \caption{A sample batch where both the positive pair ({\it green}) and the false negative pair ({\it red}) exist. The green bar and gray bar represent the similarity between all text and the positive video before and after reweighting.}
    \label{fig:reweight_example}
\end{figure}

As shown in Fig.~\ref{fig:reweight_example}, the false negative sample with similar emotion to the positive sample is greatly suppressed, while other negatives are not affected. Note that when $i\ne j$ but $s_{i}= s_{j}$, we have $w_{i,j}=\infty$, which is equivalent to removing $j$th sample from the negative pairs; when $s_{i}$ and $s_{j}$ are very different, $w_{i,j}$ is negligible; we set $w_{i,i}=0$ to not affect the true positive pair. Since $s$ is the sentiment score, the sentiment-related differences are emphasized and weighted more during training. Therefore, the proposed contrastive loss is expected to provide cleaner supervision signals for learning emotion-related representations.

% Another advantage of the proposed approach is setting $\beta$ to a large value does not make the solution trivial; we instead get
% \begin{align*}\label{eq:con_loss}
% \mathrm{Info&NCE}^*(z_v,z_t,s_t)_i\ge\\
% &-\frac{\langle z_{vi},z_{ti} \rangle}{\tau}+{\sum^N_{j=1}\frac{\langle z_{vi},z_{tj} \rangle}{\tau}-\beta\sum^N_{j=1,i\ne j}\frac{1}{\mathrm{KL}(s_{ti}\|s_{tj})}}\\
% \end{align*}
% which is very similar to a supervised loss that aligns $z_{vi}$ with $z_{ti}$ and a distillation loss that utilize the information from  $s_{ti}$.
% \begin{proof}
% \begin{align*}
% \log&{\sum^N_{j=1}\exp(\langle z_{vi},z_{tj} \rangle/\tau-w_{ij})}\\
% =&\log{\sum^N_{j=1}\frac{\exp(\langle z_{vi},z_{tj} \rangle/\tau)}{\exp(w_{ij})}}\\
% \ge&{\sum^N_{j=1}\log\frac{\exp(\langle z_{vi},z_{tj} \rangle/\tau)}{\exp(w_{ij})}}\\
% \ge&{\sum^N_{j=1}\langle z_{vi},z_{tj} \rangle/\tau-\log\exp(w_{ij})}\\
% \ge&{\sum^N_{j=1}\langle z_{vi},z_{tj} \rangle/\tau-\sum^N_{j=1}w_{ij}}\\
% \ge&{\sum^N_{j=1}\langle z_{vi},z_{tj} \rangle/\tau-\sum^N_{j=1,i\ne j}\frac{\beta}{\mathrm{KL}(s_{ti}\|s_{tj})}}\\
% \ge&{\sum^N_{j=1}\frac{\langle z_{vi},z_{tj} \rangle}{\tau}-\beta\sum^N_{j=1,i\ne j}\frac{1}{\mathrm{KL}(s_{ti}\|s_{tj})}}\\
% \end{align*}
% \end{proof}

\section{Experiments and Results}
We first introduce the datasets for evaluation in Sec.~\ref{sec:datasets} before analyzing various components of EmotionCLIP in Sec.~\ref{sec:ablation}. Then, we compare EmotionCLIP with the state-of-the-art methods on various datasets in Sec.~\ref{sec:comparison_sota}. Please see Appendix for more experimental results.

\subsection{Datasets and Evaluation Metrics}
\label{sec:datasets}
We evaluate the performance of EmotionCLIP on a variety of recently published challenging benchmarks, including four video datasets and an image dataset. The annotations of these datasets are mainly based on three physiological models: Ekman's basic emotion theory~\cite{ekman1992ekman} (7 discrete categories), the fine-grained emotion model~\cite{demszky-etal-2020-goemotions} (26 discrete categories), and the Valence-Arousal-Dominance emotion model~\cite{russell1977vad} (3 continuous dimensions). The evaluation metrics are consistent with previous methods.

% \noindent\textbf{BoLD} is a dataset for understanding human body language in the wild containing annotated short-video samples of bodily expressions of emotions. The dataset consists of 9,827 video clips and 13,239 instances, i.e., the person in these clips. Each instance is annotated with one or several of the 26 discrete categories as described in. Consistent with prior methods\cite{?}, we report the average precision (AP) and ROC-AUC on this dataset.

\noindent\textbf{BoLD}~\cite{BoLD} is a dataset for understanding human body language in the wild, consisting of 9,827 video clips and 13,239 instances, in which each instance is annotated with 26 discrete categories and VAD dimensions. 
%We report mAP and ROC-AUC for discrete labels and $R^2$ for continuous dimensions.

\noindent\textbf{MovieGraphs}~\cite{moviegraphs} is a dataset for understanding human-centric situations consisting of graph-based annotations on social events that appeared in 51 popular movies. Each graph comprises multiple types of nodes to represent actors' emotional and physical attributes, as well as their relationships and interactions. Following the preprocessing and evaluation protocol proposed in previous work~\cite{emotic, affect2mm}, we extract relevant emotion attributes from the graphs and group them into 26 discrete emotion categories. 
%We report the top-1 accuracy on this dataset. 

\noindent\textbf{MELD}~\cite{MELD} is an extension to the EmotionLines~\cite{chen2018emotionlines}, which is an emotion corpus of multi-party conversations initially proposed in the NLP domain. It offers the same dialogue examples as EmotionLines and includes audio and visual modalities along with the text. It contains around 1,400 dialogues and 13,000 utterances from the Friends tv show, where each example is annotated with 7 discrete categories. 
%We report the weighted $F_1$ score and accuracy on this dataset. 

\noindent\textbf{Liris-Accede}~\cite{Liris-Accede} is a dataset that contains videos from a set of 160 professionally made and amateur movies covering a variety of themes. Valence and arousal scores are provided continuously (\ie, every second) along movies. 
%We report the MSE on this dataset. 

\noindent\textbf{Emotic}~\cite{emotic} is an image dataset for emotion recognition in context, comprising 23,571 images of 34,320 annotated individuals in unconstrained real-world environments. Each subject is annotated with 26 discrete categories. 
%We report the mAP and ROC-AUC on this dataset.

\subsection{Ablation Study}
\label{sec:ablation}

\begin{table}[t]
\resizebox{\columnwidth}{!}{
\begin{tabular}{llll}
\toprule
% \hline
                    & mAP   & AUC   & $R^2$    \\ \midrule
EmotionCLIP (vanilla)  & 21.97 & 68.85 & 0.130 \\
\midrule
+ SAAM        & 21.53\scalebox{0.75}{\color{Maroon} $-0.44$} & 68.56\scalebox{0.75}{\color{Maroon}$-0.29$} & 0.137\scalebox{0.75}{\color{ForestGreen} $+0.007$} \\
+ SAP         & 22.28\scalebox{0.75}{\color{ForestGreen}$+0.31$} & 69.06\scalebox{0.75}{\color{ForestGreen}$+0.21$} & 0.131\scalebox{0.75}{\color{ForestGreen} $+0.001$} \\
\midrule
+ SAP \& SNCE & 22.51\scalebox{0.75}{\color{ForestGreen}$+0.54$} & 69.30\scalebox{0.75}{\color{ForestGreen}$+0.45$} &0.133\scalebox{0.75}{\color{ForestGreen} $+0.003$}\\
% \hline
\bottomrule
\end{tabular}
}
\centering
\caption{Component-wise analysis of our method on BoLD.}
\label{tab:ablation_overview}
\end{table}

\subsubsection{Analysis of Subject-Aware Context Encoding}
\label{sec:ablation_subject}

% \begin{table}[ht]
% \begin{tabular}{lccc}
% % \hline
%                                   & mAP   & AUC   & R2    \\ \hline
% Baseline                          & 21.03 & 68.16 & 10.00 \\
% \hline
% + attn mask         & 20.78 & 68.33 & 10.00 \\
% + prompting         & 21.92 & 69.03 & 10.00 \\
% \hline
% \multirow{2}{*}{+ \parbox{12em}{
% subject-aware prompting\\ 
% sentiment reweighting
% }}           & \multirow{2}{*}{22.49} & \multirow{2}{*}{69.13} & \multirow{2}{*}{10.00}\\ \\
% % \hline
% \end{tabular}
% \centering
% \caption{}
% \label{tab:ablation_overview}
% \end{table}

In this series of experiments, we start with a vanilla model and analyze it by adding various subject-aware approaches. As shown in Table~\ref{tab:ablation_overview}, decent results can be achieved in downstream tasks using the vanilla EmotionCLIP. This result supports our argument that models can learn non-trivial emotion representations from human verbal and nonverbal expressions by matching them together.

The SAP achieves better results and improves over the baseline by a reasonable margin. This improvement demonstrates the design of SAP can incorporate location-specific information to guide the model in acquiring target-related content without impacting global information modeling.

Additionally, we note that the model with SAAM yields mediocre performance. As discussed earlier, SAAM can be regarded as an efficient implementation of
the multi-stream strategy in the Transformer. This outcome suggests that the multi-stream strategy, commonly used in previous methods, may not be optimal. 
To rule out the possibility of fusion at inappropriate layers, we explore the impact of different fusion positions by applying SAAM up to a certain layer in the Transformer. It shows that the performance change does not correlate to the fusion layer change, and SAAM consistently underperforms SAP, irrespective of the fusion location. This finding implies that imposing hard masks on the model's attention may introduce unanticipated biases, while adaptively modeling context-subject interaction is more reasonable. In subsequent experiments and discussions, we use SAP as the standard implementation, unless otherwise stated.

\begin{figure}[ht] 
    \centering
    \includegraphics[width=0.9\linewidth]{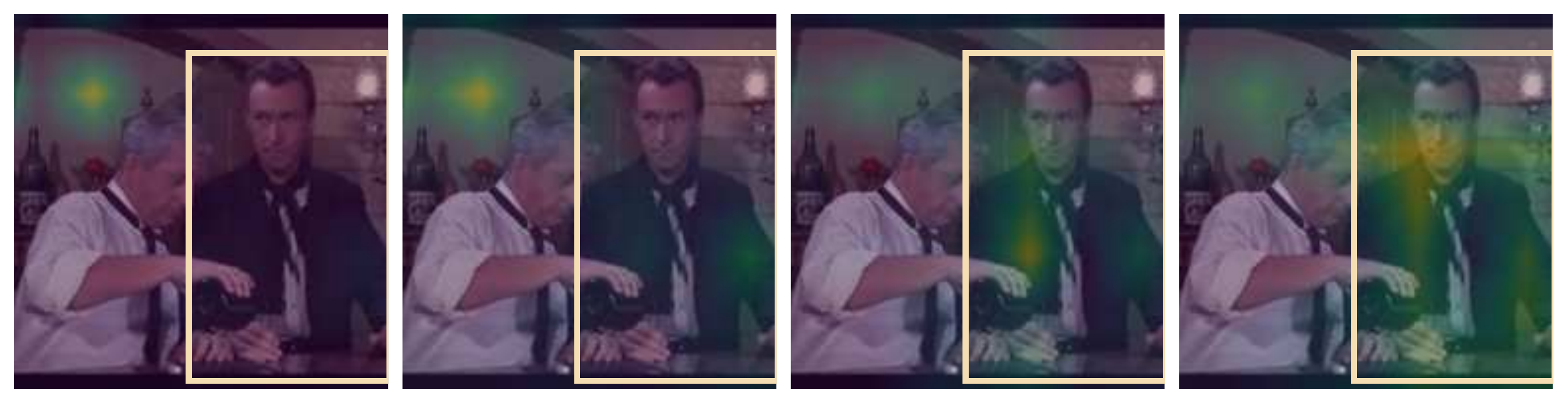}
    \includegraphics[width=0.9\linewidth]{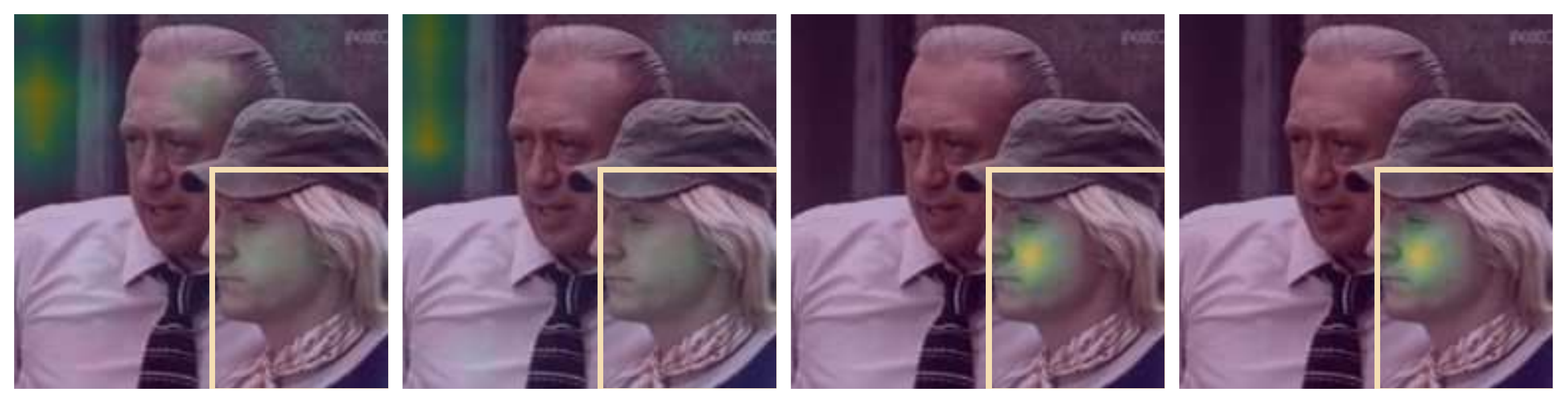}
    \includegraphics[width=0.9\linewidth]{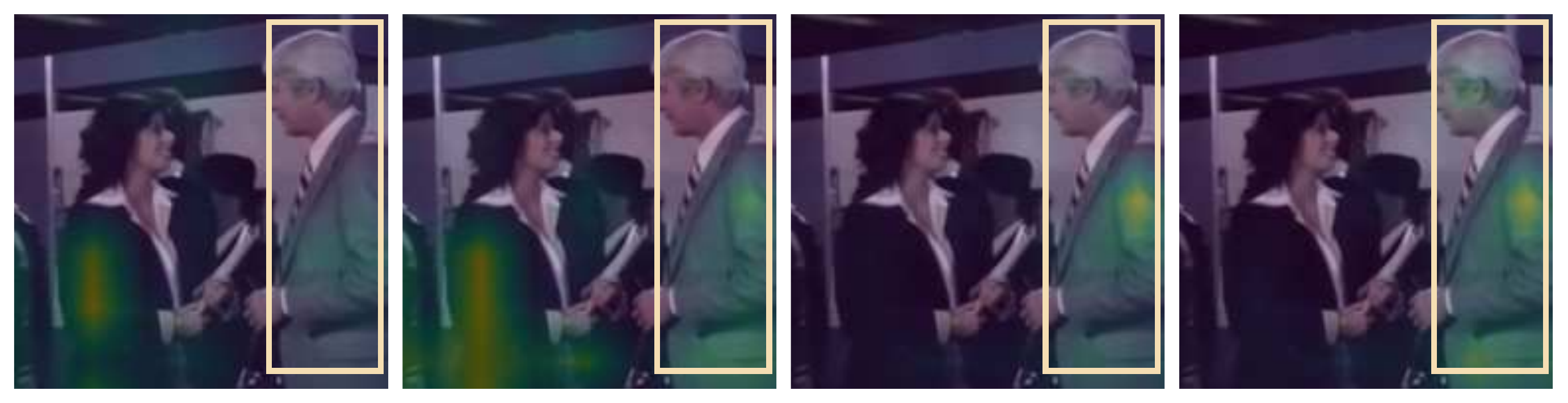}
    \caption{Attention weights for the $\mathrm{HMN}$ token from layer 1-4 ({\it left to right}) of the frame encoder in one trained network. Each row represents one frame. The green and yellow spots are the high-attention areas.}
    \label{fig:object_aware_examples}
\end{figure}

\noindent{\bf Qualitative Analysis.}
SAP offers merely a positional hint, as opposed to the mandatory attention-shifting in SAAM. Since the purpose of SAP is to ensure subject-aware encoding, it is necessary to understand if the attention guidance is appropriate. We analyze SAP by plotting $\mathrm{HMN}$ token's attention to all patches on the image. As shown in Fig.~\ref{fig:object_aware_examples}, $\mathrm{HMN}$ tokens first focus on random locations, but gradually turn their attention to the subject (\ie, the person with a bounding box) as we move to later layers, demonstrating that SAP offers sufficient guidance to the network attention.

\subsubsection{Analysis of Sentiment-Guided Contrastive Learning}
\label{sec:ablation_reweighting}
\begin{figure}[ht]
    \centering
    \includegraphics[width=\columnwidth]{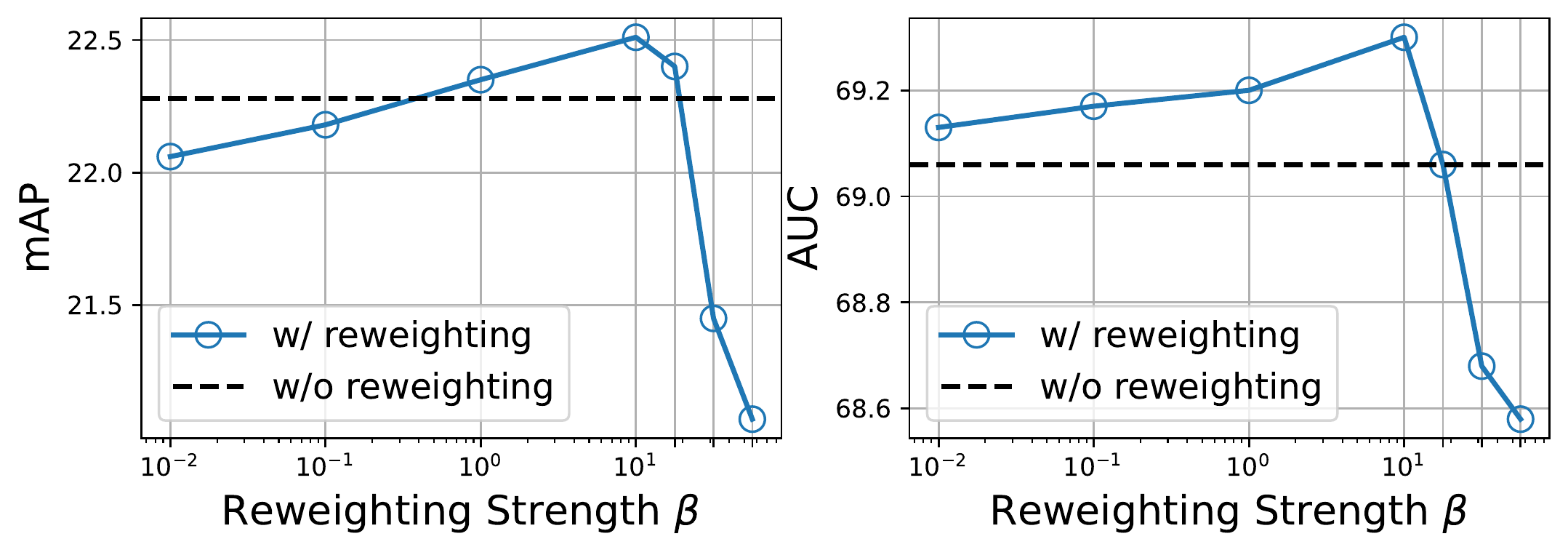}
    \caption{The effect of varying the strength of reweighting in SNCE. The larger the $\beta$, the stronger the suppression of negative samples.}
    \label{fig:ablation_reweighting_beta}
\end{figure}

We first compare models trained with different $\beta$, the hyperparameter used to control the strength of reweighting in SNCE. Note that the training objective is equivalent to the vanilla infoNCE loss~\cite{oord2018nce} when $\beta$ is set to zero. As $\beta$ increases, more negative samples within the batch are suppressed. As shown in Table~\ref{tab:ablation_overview} and Fig.~\ref{fig:ablation_reweighting_beta}, reweighting with appropriate strength can significantly increase the performance of the model as it guides the direction of learning by eliminating some significant false negatives. However, an excessively large $\beta$ can hinder the training of the model, which is within expectation. First, the sentiment scores used in the reweighting process are weak pseudo-labels provided by a pre-trained sentiment analysis model, which is not entirely reliable and accurate. Second, previous work has clearly demonstrated that batch size has a decisive impact on self-supervised learning~\cite{chen2020simclr, he2020moco, CLIP}. A too-large $\beta$ will cause too many negative samples to be suppressed, reducing the effective batch size and thus hindering the learning process. 

\begin{table}[ht]
\begin{tabular}{p{0.65\linewidth}|c}
\toprule
Text & Logit \\ \hline
I'm sorry to keep you any longer than is necessary. \textcolor{ForestGreen}{(positive sample)} & 10.65 \\ \hline
I'm sorry Tom couldn't join us. & 12.6 $\rightarrow$ \textcolor{lightgray}{-697} \\ %\hline
I hate to say it guys but it's getting late. & 9.81 $\rightarrow$ \textcolor{lightgray}{-97.6} \\ \hline
I still say it was a wild idea. & 13.8 $\rightarrow$ 13.6 \\ %\hline
Well, that was truly fascinating. & 15.2 $\rightarrow$ 15.0 \\
\bottomrule
\end{tabular}
\centering
\caption{An example batch containing both false negative and true negative samples. The logits represent the similarity between every text input and the positive video from a random epoch during training. $\rightarrow$ represents the sentiment-guided reweighting process.}\label{tab:sentiment_reweight}
\end{table}

\noindent{\bf Qualitative Analysis.}
We show how the text expressing different emotions are treated by our sentiment-guided loss. Given a positive pair, the logits are the scaled similarities between texts and the positive video; the model is penalized on large logits unless it is associated with the positive text. As shown in Table~\ref{tab:sentiment_reweight}, the texts in the second and third rows provide undesired contributions to the loss as they express similar emotions as the positive sample. After reweighting, false negatives ({\it 2nd and 3rd}) are effectively eliminated while true negatives ({\it 4th and 5th}) are negligibly affected.

% We further explore the effect of our design on scalability by varying the training batch size. Extensive work has demonstrated that larger batch sizes are beneficial for self-supervised learning. However, as shown in Figure~\ref{fig:ablation_reweighting_batchsize}, we notice that without the use of sentence-guided loss reweighting, the model starts to saturate as the batch size increases. This is because those false negatives hinder the convergence of contrastive learning-based objectives due to the appearance of contradicting objectives. We overcome this problem by false negative suppression via loss reweighting, encouraging the model to learn discriminative emotion representations by contrasting desirable negative pairs. As shown in Figure~\ref{fig:ablation_reweighting_batchsize}, our design can benefit from a larger batch size, thereby improving the scalability of the model. In subsequent experiments and discussions, we use beta=1 for the sentiment-guided loss reweighting as the default implementation unless otherwise specified.

\subsubsection{Analysis of Model Implementation}
\label{sec:ablation_model}

The frame and text encoders of our model are initialized with the image-text pre-training weights from CLIP~\cite{CLIP}. Research in neural science has demonstrated the necessity of basic visual and language understanding capabilities for learning high-level emotional semantics~\cite{pessoa2010emotion, satpute2021neural}. To validate the effectiveness of using image-text pre-training for weight initialization, we evaluate variants with different implementations. 

We first consider encoders with random initialization. As shown in Table~\ref{tab:ablation_model_implementation}, the model's performance drops sharply when training from scratch. This result is within expectation for two reasons. From an engineering perspective, previous works have demonstrated the necessity of using pre-training weights for large video models~\cite{bertasius2021timesformer, arnab2021vivit} and vision-language models~\cite{lei2021clipbert, xclip}. From a cognitive point of view, it is nearly infeasible to learn abstract concepts directly without basic comprehension skills~\cite{pessoa2010emotion}; if the model cannot recognize people, it is impossible to understand body language and facial expressions properly.

We then consider frozen encoders with pre-training weights. This is a standard paradigm for video-language understanding that trains models with offline extracted features~\cite{videoclip, li2020hero}. As shown in Table~\ref{tab:ablation_model_implementation}, our model with variants using fixed encoders performs worse compared with the model using trainable encoders. This reflects the fact that affective tasks rely on visual and verbal semantics differently from low-level vision tasks, which is what CLIP and its successors overlooked~\cite{lin2022frozen}.

We study the effect of the temporal encoder. We consider a variant where the temporal encoder is replaced by a mean pooling layer that simply averages the features of all frames. As shown in Table~\ref{tab:ablation_model_implementation}, the performance gap is obvious compared with the baseline. This phenomenon suggests that temporal dependency plays a vital role in emotion representations.

\begin{table}[t]
\resizebox{\columnwidth}{!}{
\begin{tabular}{cccll}
\toprule
\small{\makecell{Text\\Encoder}}  & \small{\makecell{Frame\\Encoder}}  & \small{\makecell{Temporal\\Encoder}} & mAP   & AUC  \\ \midrule
\cmark & \cmark & -       & 22.51 & 69.30   \\ \midrule
-      & \cmark & -       & 12.43\scalebox{0.75}{\color{Maroon} $-45\%$} & 54.96\scalebox{0.75}{\color{Maroon} $-21\%$}  \\
-      & -      & -       & 11.02\scalebox{0.75}{\color{Maroon} $-51\%$} & 50.28\scalebox{0.75}{\color{Maroon} $-27\%$}  \\ \midrule
\xmark & \cmark & -       & 13.40\scalebox{0.75}{\color{Maroon} $-40\%$} & 57.38\scalebox{0.75}{\color{Maroon} $-17\%$}  \\
\xmark & \xmark & -       & 18.43\scalebox{0.75}{\color{Maroon} $-18\%$} & 65.08\scalebox{0.75}{\color{Maroon} $-6.0\%$}  \\ \midrule
\cmark & \cmark & $\circ$ & 21.17\scalebox{0.75}{\color{Maroon} $-5.9\%$} & 68.74\scalebox{0.7}{\color{Maroon} $-0.8\%$}  \\ 
\bottomrule
\end{tabular}
}
\centering
\caption{Ablation study on different model implementations. \cmark means trainable, initialization with pre-training weights. \xmark means frozen, initialization with pre-training weights. - means trainable, random initialization. $\circ$ means no parameters.}
\label{tab:ablation_model_implementation}
\end{table}

% \subsubsection{Analysis of Training Data}
% \label{sec:ablation_dataset}
% \begin{figure}[ht]
%     \centering
%     \includegraphics[width=0.45\linewidth]{example-image}
%     \includegraphics[width=0.45\linewidth]{example-image}
%     \caption{Caption}
%     \label{fig:ablation_filtering_threshold}
% \end{figure}

% - quality and quantity (neutral-score filtering threshold)
% - scalability (number of samples)

% MAIN TABLE
\begin{table*}[t!]
\label{tab:comparison_all}
\centering

\footnotesize{
\begin{subtable}[t]{0.345\textwidth}\centering
% BOLD
\caption{BoLD~\cite{BoLD}}
\begin{tabular}{lccc}
\toprule
Method      & mAP $\uparrow$ & AUC $\uparrow$ & $R^2\uparrow$ \\ \midrule
\multicolumn{4}{l}{\textit{Supervised}}    \\
ST-GCN~\cite{STGCN}      &  12.63   &  55.96   & 0.044   \\
TSN~\cite{TSN}         &  17.02   &  62.70   & 0.095   \\
Filntisis~\etal~\cite{Filntisis2020bold}   &  16.56   &  62.66   & 0.092   \\
Pikoulis~\etal~\cite{Pikoulis2021LeveragingSS}    &  19.29   &  66.82   & {\bf 0.149}   \\ 
\midrule
\multicolumn{4}{l}{\textit{Linear-Eval}} \\
VideoCLIP~\cite{videoclip}   &   11.19  &   51.23  & -5.11   \\
X-CLIP~\cite{xclip}      &   13.26  &   56.86  &  -0.03   \\
\rowcolor{blue!10}
EmotionCLIP    &   {\bf 22.51}  &   {\bf 69.30}  & 0.133   \\
\bottomrule
\end{tabular}
\label{tab:comparison_bold}

\vskip 0.8cm
% MovieGraphs
\caption{MovieGraphs~\cite{moviegraphs}}
\begin{tabular}{p{0.40\linewidth}cc}
\toprule
Method      & \makecell{Val Acc $\uparrow$} & \makecell{Test Acc $\uparrow$} \\ 
\midrule
\multicolumn{3}{l}{\textit{Supervised}}    \\
EmotionNet~\cite{emotionnet}  & 35.60  & 27.90    \\
*Affect2MM~\cite{affect2mm}  & 39.88  & 30.58    \\ \midrule
\multicolumn{3}{l}{\textit{Linear-Eval}} \\
VideoCLIP~\cite{videoclip}   & 29.91  & 23.44    \\
X-CLIP~\cite{xclip}      & 29.06      & 23.58    \\
\rowcolor{blue!10}
EmotionCLIP          & {\bf 41.60}  & {\bf 32.35}    \\
\bottomrule
\end{tabular}
\label{tab:comparison_mg}
\end{subtable}
}
\hfill
% MELD
\footnotesize{
\begin{subtable}[t]{0.31\textwidth}\centering
% Emotic
\caption{Emotic~\cite{emotic}}
\centering
\begin{tabular}{p{0.465\linewidth}cc}
\toprule
Method                & mAP $\uparrow$ & AUC $\uparrow$ \\ \midrule
\multicolumn{3}{l}{\textit{Supervised}}    \\
CAER-Net~\cite{CAER}              & 20.84  &  -  \\
Affective Graph~\cite{affective_graph}       & 28.42  &  -  \\
Fusion Model~\cite{emotic}          & 29.45  &  -  \\
EmotiCon \scalebox{0.75}{(GCN)}~\cite{emoticon}        & 32.03  &  -  \\
EmotiCon \scalebox{0.75}{(Depth)}~\cite{emoticon}      & {\bf 35.48}  & -  \\ \midrule
\multicolumn{3}{l}{\textit{Linear-Eval}} \\
% CLIP~\cite{CLIP}                  & 33.21  &  71.18  \\
VideoCLIP~\cite{videoclip}             & 19.92      &  56.31  \\
X-CLIP~\cite{xclip}                & 22.80      &  61.31 \\
\rowcolor{blue!10}
EmotionCLIP                    & 32.91  &  {\bf 71.41}  \\
\bottomrule
\end{tabular}
\label{tab:comparison_emotic}

% MELD
\vskip 0.475cm
\caption{MELD~\cite{MELD}}
\centering
% \begin{tabular}{p{0.46\linewidth}cc}
% \toprule
% Method          & W. $F_1 \uparrow$ & Acc $\uparrow$ \\ \midrule
% \multicolumn{3}{l}{\textit{Supervised}}    \\
% M2FNet \scalebox{0.75}{(Visual)}~\cite{m2fnet} & 32.44       & 45.63   \\
% *M2FNet~\cite{m2fnet}          & 66.71       & 67.85   \\ 
% \midrule
% \multicolumn{3}{l}{\textit{Linear-Eval}} \\
% VideoCLIP~\cite{videoclip}       & 32.06           & 45.19   \\
% X-CLIP~\cite{xclip}          & 32.46          & 38.31   \\
% \rowcolor{blue!10}
% EmotionCLIP            & 34.93           & 48.70   \\
% \bottomrule
% \end{tabular}
\begin{tabular}{p{0.465\linewidth}cc}
\toprule
Method          & Acc $\uparrow$ & W. $F_1 \uparrow$ \\ \midrule
\multicolumn{3}{l}{\textit{Supervised}}    \\
M2FNet \scalebox{0.75}{(Visual)}~\cite{m2fnet} &   45.63 &   32.44 \\
*M2FNet~\cite{m2fnet}          & 67.85       & 66.71   \\ 
\midrule
\multicolumn{3}{l}{\textit{Linear-Eval}} \\
VideoCLIP~\cite{videoclip}       & 45.19           & 32.06   \\
X-CLIP~\cite{xclip}          &  38.31         &  32.46   \\
\rowcolor{blue!10}
EmotionCLIP            &  {\bf 48.28}      & {\bf 34.59}   \\
\bottomrule
\end{tabular}
\label{tab:comparison_meld}

\end{subtable}
}
\hfill
\footnotesize{
\begin{subtable}[t]{0.32\textwidth}
\centering
% Liris-Accede
\caption{Liris-Accede~\cite{Liris-Accede}}
\begin{tabular}{lcc}
\toprule
Method      & \makecell{V. MSE $\downarrow$} & \makecell{A. MSE $\downarrow$} \\ \midrule
\multicolumn{3}{l}{\textit{Supervised}}    \\
Quan~\etal~\cite{quan2018frame}        & 0.115           & 0.171           \\
Ko~\etal~\cite{ko2018towards}          & 0.102           & {\bf 0.149}           \\
*CERTH-ITI~\cite{batziou2018visual}  & 0.117           & 0.138           \\
*THUHCSI~\cite{jin2017thuhcsi}    & 0.092           & 0.140           \\
*Yi~\etal~\cite{yi2018cnn}         & 0.090           & 0.136           \\
*GLA~\cite{sun2019gla}        & 0.084           & 0.133           \\
*Zhao~\etal~\cite{zhao2019video}       & 0.071           & 0.137           \\
*Affect2MM~\cite{affect2mm}  & 0.068           & 0.128           \\ 
\midrule
\multicolumn{3}{l}{\textit{Linear-Eval}} \\
VideoCLIP~\cite{videoclip}   & 0.142               & 0.151           \\
X-CLIP~\cite{xclip}      & 0.133              & 0.246          \\
\rowcolor{blue!10}
EmotionCLIP   & {\bf 0.096}    & 0.155          \\
\bottomrule
\end{tabular}
\label{tab:comparison_liris}
\vskip 0.42cm
\begin{tabular}{p{0.42\linewidth}<{\centering}p{0.42\linewidth}<{\centering}}
\toprule
Abbreviation      & Meaning \\ \midrule
A. MSE   & Arousal MSE              \\
V. MSE      & Valence MSE                    \\
W. $F_1$        & Weighted $F_1$                 \\
Acc.        & Top-1 Accuracy                \\
$\downarrow$       & Lower is better               \\
$\uparrow$     & Higher is better               \\
\bottomrule
\end{tabular}
\end{subtable}}

\caption{
Comparisons to the state-of-the-art across multiple datasets. 
Methods marked with * use multimodal inputs, \ie, audio and text.
Bold numbers indicate the best results achieved using visual inputs only.
}
\end{table*}
% END OF MAIN TABLE

\subsection{Comparison with the State of the Art}
\label{sec:comparison_sota}

Based on our previous ablation experiments, we choose the model with SAP and SNCE as the default implementation and compare it with the state-of-the-art. In addition, we also compare with VideoCLIP~\cite{videoclip} and X-CLIP~\cite{xclip}, both of which are state-of-the-art vision-language pre-training models for general video recognition purposes. To evaluate the quality of learned representations, we follow the practice in CLIP~\cite{CLIP} and use linear-probe evaluation protocol~\cite{chen2020simclr, he2020moco} for vision-language pre-training models.

\noindent\textbf{BoLD.}
As shown in Table~\ref{tab:comparison_bold}, EmotionCLIP substantially outperforms the state-of-the-art supervised learning methods on the challenging 26-class emotion classification task and achieves comparable results on continuous emotion regression. It is worth noting that a complex multi-stream model is used in \cite{Pikoulis2021LeveragingSS} to integrate the human body and context information, while we achieve better results with a single-stream structure using RGB information only. This difference reflects that the subject-aware approach we designed models the relationship between the subject and context.
We also notice that other vision-language-based methods perform poorly on emotion recognition tasks, although they are designed for general video understanding purposes. This phenomenon is largely attributed to the lack of proper guidance; the model can only learn low-level visual patterns and fails to capture semantic and emotional information.

\noindent\textbf{MovieGraphs.}
As shown in Table~\ref{tab:comparison_mg}, EmotionCLIP substantially outperforms the best vision-based method and even surpasses Affect2MM~\cite{affect2mm}, a powerful multimodal approach that uses audio and text descriptions in addition to visual information. Instead, other vision-language pre-training models are still far from supervised methods.

\noindent\textbf{MELD.}
EmotionCLIP performs well on MELD as shown in Table~\ref{tab:comparison_meld}; it achieves comparable results to the state-of-the-art vision-based methods. It is worth noting that this dataset is extended from an NLP dataset, so the visual data is noisier than the original text data. In fact, according to the ablation experiments in~\cite{m2fnet}, it is possible to achieve an accuracy of 67.24\% using only text, while adding visual modality information only improves the accuracy by about 0.5\%. This result explains why our method significantly lags behind multimodal methods using text inputs.

\noindent\textbf{Liris-Accede.}
As shown in Table~\ref{tab:comparison_liris}, EmotionCLIP achieves promising results using visual inputs only. It even competes with many multimodal approaches that are benefited from the use of audio features~\cite{yoon2018multimodal}.

\noindent\textbf{Emotic.}
As shown in Table~\ref{tab:comparison_emotic}, EmotionCLIP outperforms all RGB-based supervised methods while other vision-language models perform poorly. The improvement of~\cite{emoticon} is attributable to the use of additional depth information. This result demonstrates the capability of EmotionCLIP in learning relevant features from complex environments.

\section{Conclusion}
% We present EmotionCLIP, the first pre-training paradigm for visual emotion understanding. Our pre-training method only utilizes uncurated data, significantly reducing the need for manually labeled datasets. Comprehensive ablation studies reveal that the proposed verbal and nonverbal guidance further enhances the learned features. EmotionCLIP outperforms many supervised multimodal state-of-the-art methods despite only using linear evaluation protocol. 

The pre-training methodology, which has brought about significant advancements in numerous CV and NLP domains, has not yet been employed in AEI research. We address this void by introducing EmotionCLIP, the first vision-language pre-training framework that circumvents the need for curated data and annotations. Our study establishes the viability of acquiring generalized emotion representations directly from human communication, thereby considerably broadening the horizons of current AEI research. The emergence of this pre-training paradigm offers an alternative solution to data scarcity, paving the way for a myriad of potential applications. We anticipate that our work will stimulate further investigation in this area and contribute to the evolution of more adaptable and efficacious approaches for affective computing.

\noindent\paragraph{Acknowledgments.} This research was supported by generous gifts from the Amazon Research Awards program. The work used computational resources 
from the Advanced Cyberinfrastructure Coordination Ecosystem: Services \& Support (ACCESS) program, which is supported by National Science Foundation.
%grants 2138259, 2138286, 2138307, 2137603, and 2138296.
%from the Extreme Science and Engineering Discovery Environment (XSEDE), which is supported by National Science Foundation grant No. ACI-1548562.

%%%%%%%%% REFERENCES
{\small
\bibliographystyle{ieee_fullname}
\bibliography{main}
}

\end{document}